\documentclass[a4paper, 10pt, conference]{ieeeconf} 
\IEEEoverridecommandlockouts                              % This command is only needed if 
\usepackage[utf8x]{inputenc}

\usepackage{changes}%用于撰写过程中的修改与注释
	\definechangesauthor[name={Sanqing Qu}, color=orange]{per}
	\setremarkmarkup{(#2)}

\usepackage{float}
\usepackage[caption = true]{subfig}
\usepackage{graphicx}
\graphicspath{{figures/}} %Setting the graphicspath   
\usepackage{algorithm}  
\usepackage{algorithmicx}
\usepackage{algpseudocode}
\usepackage{amsmath}
\usepackage{amssymb}%空集符号
\usepackage{booktabs}
\floatname{algorithm}{Algorithm}

\title{\LARGE \bf
	Self-Localization of Parking Robots Using Square-Like Landmarks
}

\author{Canbo Ye$^{1}$% <-this % stops a space	
	\thanks{$^{1}$Canbo Ye is with School of Automotive Studies, 
		Tongji University,Shanghai,China,
		{
			\tt\small albert.yecanbo@outlook.com}
		}%
}

\author{Canbo Ye$^{1}$, Guang Chen$^{1,2,*}$,Sanqing Qu$^{1}$, Qianyi Yang$^{1}$, Kai Chen$^{1}$, Jiatong Du$^{1}$, Ruien Hu$^{1}$	
	\thanks{${1}$ College of Automotive Engineering, Tongji University}
	\thanks{${2}$ Chair of Robotics, Artificial Intelligence and Real-time Systems TUM Department of Informatics Technical University of Munich }\\
	\thanks{${*}$ Corresponding author E-mail address: {\tt\small guang@in.tum.de}}
}

\begin{document}

	\maketitle
	\thispagestyle{empty}
	\pagestyle{empty}
	
\begin{abstract}
In this paper, we present a framework for self-localization of parking robots in a parking lot innovatively using square-like landmarks, aiming to provide a positioning solution with low cost but high accuracy. It utilizes square structures common in parking lots such as pillars, corners or charging piles as robust landmarks and deduces the global pose of the robot in conjunction with an off-line map. The localization is performed in real-time via Particle Filter using a single line scanning LiDAR as main sensor, an odometry as secondary information sources. The system has been tested in a simulation environment built in V-REP, the result of which demonstrates its positioning accuracy below 0.20 m and a corresponding heading error below 1$^{\circ}$.
\end{abstract}

\section{Introduction}
\begin{figure}
	\centering
	\subfloat{\includegraphics[width= 3.3in]{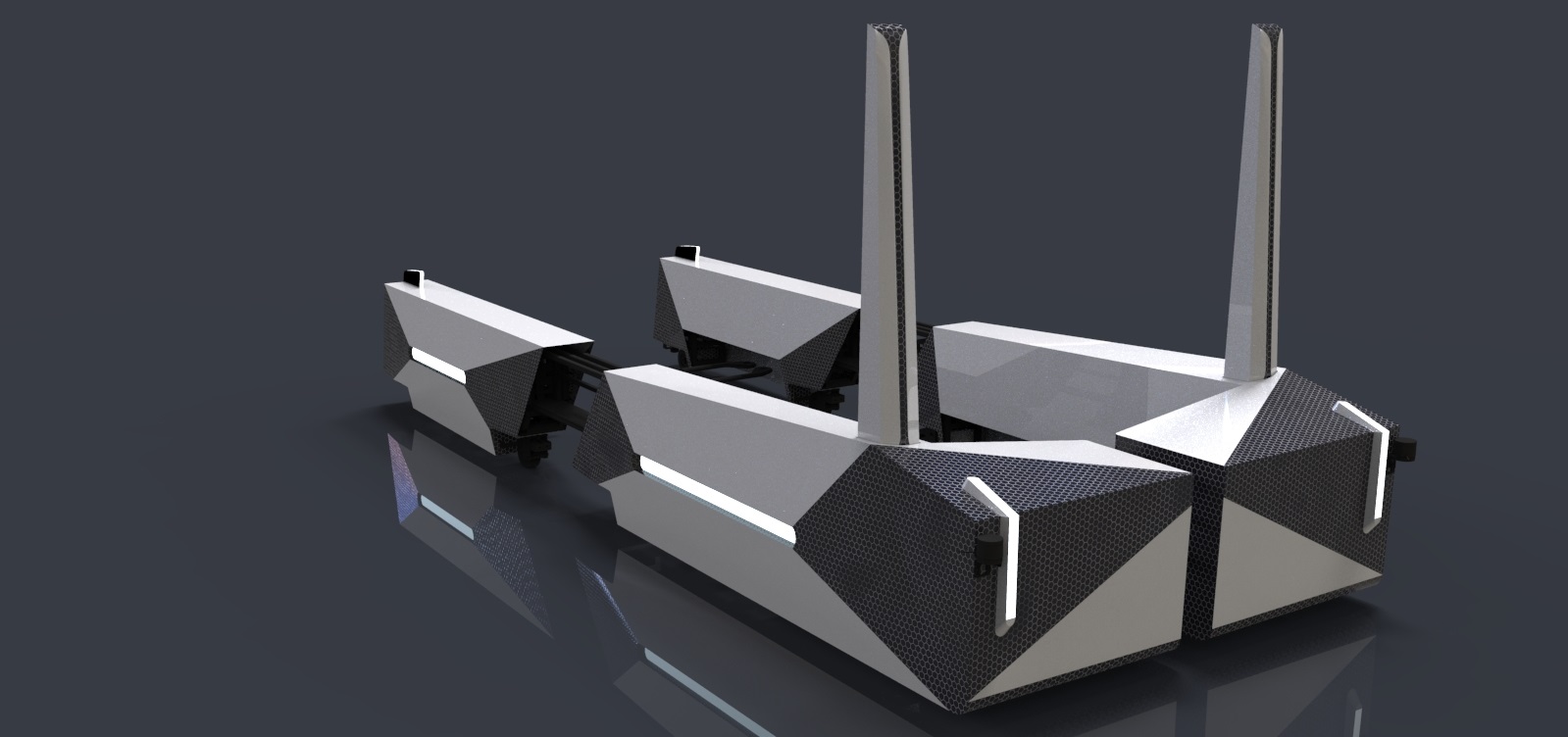}}
	\caption{The design sketch of Tongji autonomous parking robot car}
	\label{figs:robot}
\end{figure}

With the increasing number of vehicles and the shortage of parking spaces, the traditional parking lot is forced to be upgraded. In addition to the existing mechanical three-dimensional parking lot, the fully autonomous parking robot also plays a fundamental role. Our contribution to this trend is the setup of a new electric X-by-wire parking robot car, characterized by comprising two lateral spreading device deployment apparatus and a transverse, with the aim of adapting to the wheelbase and track of different vehicles. The parking robot is designed to carry vehicles from one position to a given spot without human intervention, with capability of environment perception, self-localization and path planning.  The design sketch of our autonomous parking robot is shown in Fig~\ref{figs:robot}. 

In indoor parking lots, however, it is not a straightforward task for robot to localize itself due to the absence of GPS caused by signal attenuation through construction materials, especially when high precision localization is needed. The existing approaches for self-localization indoors include techniques based on WiFi, Radio Frequency Identification Device (RFID), Ultra Wideband (UWB), Bluetooth, etc~\cite{0}. Most of these localization approaches suffer from defect of high cost, unstableness or low precision. Here we adopt a novel localization solution using square-like landmarks detected by a single line scanning LiDAR, the advantages of which lie in its low cost but high precision. Moreover, it is also workable in outdoor parking lots as long as the required landmarks are sufficient. For example, outdoor charging piles can be designed as available landmarks for our work. 

This paper is organized as follows: After discussing related works in Section II, an overview about the presented localization approach is given in Section III. The following Section IV and V focus on the detailed implementation of the approach. Then a simulation test and its results are illustrated in Section V. A conclusion in the last Section ends this work.

\section{Related Work}

In the context of landmark-based self-localization for mobile robot, different techniques have been proposed in the literature. The main task of a landmark-based localization is to collect and extract distinct features from either artificial or natural landmarks. Since the focus of our work primarily lies on the localization in a specific area --- a parking lot, we can further narrow down the scope of the literature review by concentrating mainly on strategies that is practicable in this case.

Vision based localization methods are given in many works. Among these methods, the natural landmark-based approach suffers limitations due to its sensitivity to variable environmental conditions and requirement for high computational power for image processing~\cite{1}. Compared with natural landmarks, artificial landmarks tend to be easier and faster to be sensed and recognized with a better resistance to noise. In a parking lot, QR-code is likely to be an applicable artificial landmark. In papers~\cite{2}\cite{3}\cite{4} ,  localization approach using QR-code is proposed and tested. Ceiling landmarks~\cite{5}\cite{6} are also workable in this situation. These approaches, however, require relatively more effort to set up the artificial landmarks, which is not an easy task in a large parking lot. Moreover, image landmarks including QR-Code or celling mark tend to get blurry gradually due to dust or erosion, resulting in increasing error or even failure of the localization.

For LiDAR-based localization techniques in this context, pole-like landmark~\cite{7}\cite{8} is a suitable option, but it is not common in indoor parking lots, which limits its application to an outdoor place surrounded by a certain number of pole items like trees, street lamps or traffic signs. Another shortcoming of such approach is that it needs to use multi-line LiDAR to distinguish the pole items from others and extract their geometric features, considering that horizontal size of the lamps or signs is generally small and the irregular geometric features of trees limit its positioning accuracy. 

In our localization approach based on square landmarks, several presented methods are applied. In order to represent a square landmark with four corners, we refer to the L-Shape Fitting method proposed in~\cite{CMU_LShape}. The following position estimate stage employs Particle Filter (also known as Monte Carlo Localization)~\cite{9}.

\section{overview}
\begin{figure}
	\centering
	\subfloat{\includegraphics[width = 3.2in]{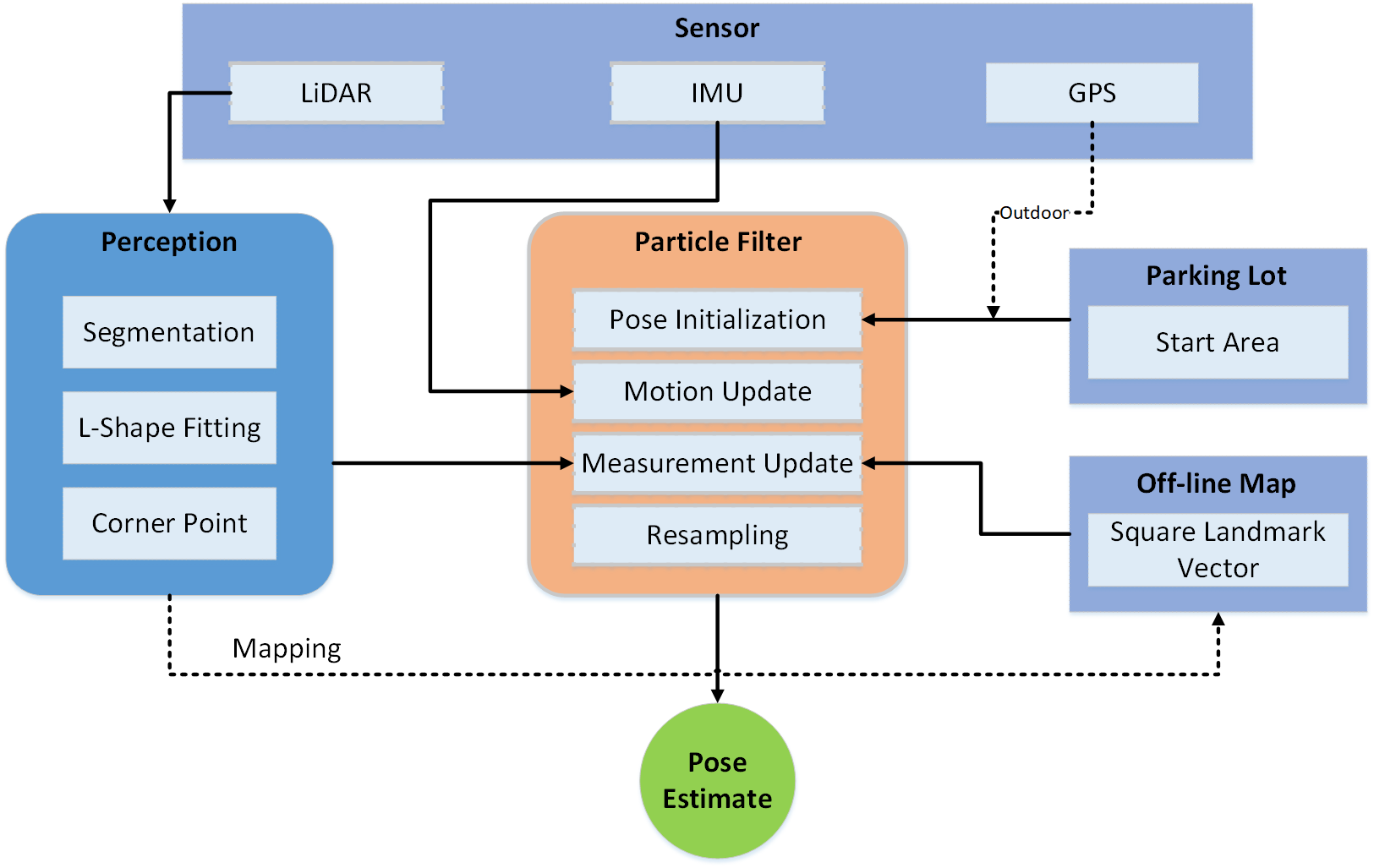}}
	\captionsetup{font={small}}
	\caption{Overview of localization framework}
	\label{figs:overview}
\end{figure}

 An overview of the developed localization framework is illustrated in Fig.~\ref{figs:overview}. The central component of the framework is Particle Filter, which utilizes and fuses the data from different sensors to deduce a continuously updated estimate of the robot pose. The pose initialization of the particles is accomplished by a GPS or a start area inside the parking lot, and IMU serves as information source of motion update. 
 
 In the measurement update step, we apply a novel landmark---square-like structures---to associate with a corresponding off-line map to estimate the current pose of the robot. The square-like structures, including square pillars and charging piles, are common in parking lots. The wall corners inside the parking lot can also fulfil the requirements of such landmarks. Moreover, It is also feasible to artificially arrange some square pillars in the parking lot if necessary. These square-like structures enjoy the advantages of being time-invariant, robustly redetectable and having distinct characteristics, leading to its usability as a reliable landmark for self-localization. 
 
 In this context, a single line scanning LiDAR is employed to perceive and extract the square landmark from surroundings. Compared with the multiline LiDAR commonly used on current smart vehicles or robots, the single line LiDAR enjoys much lower cost and better reliability with higher scanning speed and angular resolution. Although it can only perceive a single layer of 3-D information, it is fully applicable for the detection of square-like landmarks, the needed feature of which lies in the corners of its cross section. To complete this task, an L-Shape Fitting approach~\cite{CMU_LShape} is utilized, which will deduce the coordinates of the four corner points for each detected square item. Furthermore, these produced corner coordinates will be filtered and used in the measurement update of Particle Filter, associating with an off-line map containing the ground truth coordinates of the landmarks.  
 
 In order to verify the effectiveness of our approach, a simulation environment based on V-REP was built and the result of which will be shown in section VI.

\section{Square-like Landmarks extraction}
In order to utilize specific square-like structures as landmarks for Localization, other irrelevant objects, such as vehicles or some temporarily stacked square items, should be avoided in case of distraction. For this purpose, the single line scanning LiDAR is installed at a higher position of the parking robot, 2 meters above the ground. In this case, most vehicles in a parking lot will not be detected by the LiDAR. After acquiring the scanning data of the environmental objects, segmentation is firstly employed, separating data points into different clusters. Afterwards, we follow the L-shape fitting approach proposed in~\cite{CMU_LShape}, the consequence of which represents all eligible point cloud shape with a rectangle. To apply this result for localization, we calculate the corners of all deduced rectangles. These corner coordinates can be utilized in both mapping and real-time positioning. 
	\subsection{Segmentation}
	The laser scan data need to be segmented into different clusters before performing L-shape fitting. There are several classical clustering algorithms capable of this segmentation work. In~\cite{CMU_LShape}, an adaptive segmentation algorithm is employed, however, this algorithm is likely to result in an increase of noise clusters, because isolated points are not excluded. 
	
	For accurate segmentation to the range points, we adopt the DBSCAN algorithm (Density Based Spatial Clustering of Applications with Noise)~\cite{10}, the basic idea of which is to divide the range points into different clusters based on a preassigned distance threshold. The input for the segmentation algorithm is the 2-D coordinates of n range scanning points, $D\in R_{n\times2}$, which are relative to the pose of the LiDAR. The output of the algorithm is a set of segmented clusters, each of which potentially corresponds to an object in the real world. 
	
\begin{figure}
		\centering
		\subfloat{\includegraphics[width= 3.3in]{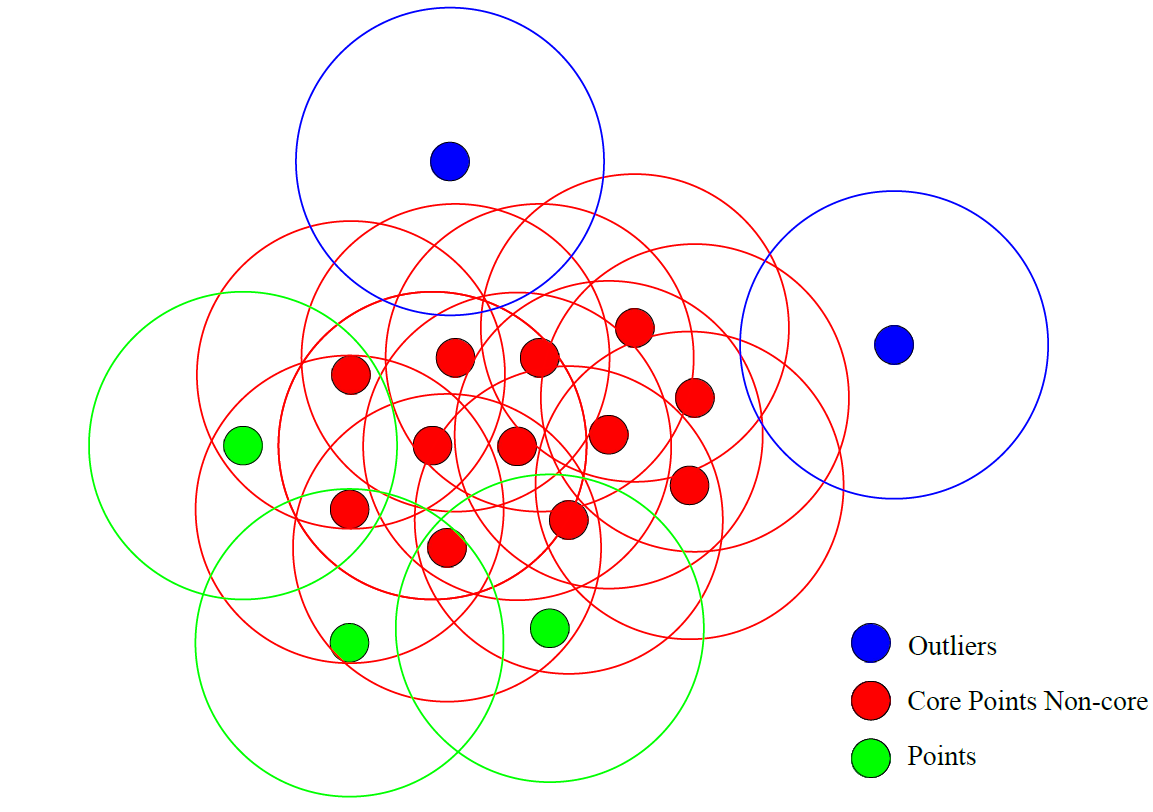}}
		\caption{Segmented points using DBSCAN (MinPts = 4) (Best viewed in color)}
		\label{figs:DBSCAN}
	\end{figure}

	The produced clusters of DBSCAN consist of core points and non-core points. The core point means that there are at least $MinPts$ points within distance threshold, while the non-core point signifies that this point is directly reachable from a core point within distance threshold. Points unreachable from any core points are outliers. In the Fig.~\ref{figs:DBSCAN}, the red points belong to core points and the green points are non-core points, while the blue points are outliers.
\iffalse
	\begin{algorithm}[t] 
		\label{Args:DBSCAN}
		\caption{DBSACN segmentation algorithm}
		\begin{algorithmic}[1]
			\Require{range data points $D\in R_{n\times2}$}
			\Ensure{set of point clusters $S = \{S_1,S_2,\cdots,S_k\}$}
			\State \mbox{init the core points group:$\Omega = \O$}
			\For {j=1,2,$\cdots$,m}
			\State \mbox{get the $Neirhbors(x_j)$ of the point $x_j$ within $r$ }
			\If{$|Neirhbors(x_j)|\geqslant MinPts$} 
			\State \mbox{insert $N_j$ into $\Omega$}
			\EndIf
			\EndFor
			\State init the number of clusters $k=0$
			\State \mbox{init the unvisited points group $\Gamma = D$}
			\While {$\Omega \neq \O$}
			\State \mbox{record current unvisited points group $\Gamma_{old} = \Gamma$}
			\State \mbox{randomly select a core point $\alpha \in \Omega$,init queue $Q = \langle \alpha \rangle$}
			\State \mbox{$\Gamma = \Gamma\setminus\{\alpha\}$}
			\While {$Q \neq \O$}
			\State \mbox{get the first sample $q$ of the queue $Q$ }
			\If \mbox{$|Neirhbors(q)|\geqslant MinPts $}
			\State \mbox{set $\Delta = Neighbors(q) \cap \Gamma$ }
			\State \mbox{insert $\Delta$ into queue $Q$}
			\State \mbox{$\Gamma = \Gamma\setminus\Delta$} 
			\EndIf
			\EndWhile
			\State \mbox{$k=k+1,S_k = \Gamma_{old}\setminus\Gamma$}
			\State \mbox{$\Omega = \Omega\setminus S_k$}	
			\EndWhile	
			\State \Return $S$	
		\end{algorithmic}
	\end{algorithm} 
	\fi
	The main procedure is that we first detect the core points of range data, and then from these core points explore the non-core points. Note that this segmentation algorithm is adaptive due to the proportional distance threshold $r$ between lidar and objects. This is justified by the underlying fact that the laser scanning longitude resolution grows with the range distance~\cite{CMU_LShape}. Furthermore, if the scanning sequence is available, points' Neighbors within the threshold distance $r$ can be more efficiently found. And a graph-based index structure can be used for speeding up neighbor search operations~\cite{DBSCAN_Speedup}.
		
	\subsection{L-Shape Fitting}
	In order to acquire the four corners of given square point clouds, a rectangle that fit best to the clusters should be found. In~\cite{CMU_LShape}, several L-Shape Fitting methods are proposed and compared, from which we apply the Seach-Based Rectangle Fitting using closeness criterion. Here is a brief illustration of the approach. 
	
	The basic idea is that all the possible directions of the rectangle are iterated and a rectangle containing all the scan points is found. Thereafter we can obtain the distances of all the points to the rectangle's four edges. The pseudocode of the algorithm is shown in Alg.~\ref{Args:shape_fitting}.
	
	By using the closeness criterion, emphasis is laid on how close the points are to the two edges of the right corner. In the projected 2-D plane, we can find $c_1^{max}$ and $c_1^{min}$ which specify the boundaries in one direction for all points, and distance vectors $c_1^{max} - C_1$ and $ C_1 - c_1^{min}$ record all the points' distance to the two boundaries, so that a closer boundary as well as the corresponding distance vector $\vec{D}_1=[d_1,...,d_m]$ can be confirmed. The distance vector $\vec{D}_2$ of the other direction is defined in the same way. Then we calculate the closeness score, which is defined as $\sum_{i=1}^m 1/d_i$, with $d_i$ being the $i^{th}$ point's distance to its closer boundary. By maximizing this score function, we finally work out the best square fitting based on the closeness criterion.
	\iffalse
	\begin{algorithm}[t]
		\caption{Closeness Criterion}
		\label{Args:Closeness Criterion}
		\begin{algorithmic}[1]
			\State \textbf{function} CalculateCloseness($C_1,C_2$)
			\State \qquad $c_1^{max} = max(C_1)$, $c_1^{min} = min(C_1)$
			\State \qquad $c_2^{max} = max(C_2)$, $c_2^{min} = min(C_2)$
			
			\Require{cluster data points $X\in R^{n\times2}$; }
			\Ensure{ rectangle edges\{$a_ix + b_iy = c_i | i = 1,2,3,4$\}}
			\State {$Q=\O$}
			\For {$\theta = 0\ to \ \pi/2 $} 
			\State $\vec{e}_1 = (cos\theta,sin\theta)$
			\State $\vec{e}_2 = (-sin\theta,cos\theta)$
			\State $\vec{X} = (x,y)$
			\State $C1 = X \cdot \vec{e}_1^T$
			\State $C2 = X \cdot \vec{e}_2^T$
			\State $q = CalculateCriterion(C1,C2)$
			\State insert $q$ into $Q$ with key($\theta$)
			\EndFor
			\State select key($\theta^*$) from $Q$ with maximum value
			\State $C_1^* = \vec{X} \cdot (cos\theta^*,sin\theta^*)$
			\State $C_2^* = \vec{X} \cdot (-sin\theta^*,cos\theta^*)$
			\State $a_1 = cos\theta_1,\quad b_1 = sin\theta_1,\quad c_1 = min\{C_1\}$
			\State $a_2 = cos\theta_2,\quad b_2 = sin\theta_2,\quad c_2 = min\{C_2\}$
			\State $a_3 = cos\theta_1,\quad b_3 = sin\theta_1,\quad c_3 = max\{C_1\}$
			\State $a_4 = cos\theta_2,\quad b_4 = sin\theta_2,\quad c_4 = max\{C_2\}$
		\end{algorithmic}
	\end{algorithm}
\fi
	\begin{algorithm}[t]
		\caption{Rectangle Fitting}
		\label{Args:shape_fitting}
		\begin{algorithmic}[1]
			\Require{cluster data points $X\in R^{n\times2}$; }
			\Ensure{ rectangle edges\{$a_ix + b_iy = c_i | i = 1,2,3,4$\}}
			\State {$Q=\O$}
			\For {$\theta = 0\ to \ \pi/2 $} 
			\State $\vec{e}_1 = (cos\theta,sin\theta)$
			\State $\vec{e}_2 = (-sin\theta,cos\theta)$
			\State $\vec{X} = (x,y)$
			\State $C1 = X \cdot \vec{e}_1^T$
			\State $C2 = X \cdot \vec{e}_2^T$
			\State $q = CalculateCloseness(C1,C2)$
			\State insert $q$ into $Q$ with key($\theta$)
			\EndFor
			\State select key($\theta^*$) from $Q$ with maximum value
			\State $C_1^* = \vec{X} \cdot (cos\theta^*,sin\theta^*)$
			\State $C_2^* = \vec{X} \cdot (-sin\theta^*,cos\theta^*)$
			\State $a_1 = cos\theta_1,\quad b_1 = sin\theta_1,\quad c_1 = min\{C_1\}$
			\State $a_2 = cos\theta_2,\quad b_2 = sin\theta_2,\quad c_2 = min\{C_2\}$
			\State $a_3 = cos\theta_1,\quad b_3 = sin\theta_1,\quad c_3 = max\{C_1\}$
			\State $a_4 = cos\theta_2,\quad b_4 = sin\theta_2,\quad c_4 = max\{C_2\}$
		\end{algorithmic}
	\end{algorithm}

	\subsection{Corner Points and Landmark Map}
	After fitting square clusters into rectangles, we acquire four lines $a_ix+b_iy=c_i |_{i=1,2,3,4}$. By calculating their intersection, we obtain four coordinates $(x_i,y_i) |_{i=1,2,3,4}$, which will be directly utilized in localization stage. These corners are stored in a digital map in form of vector. Moreover, the perception of square structures in real-time is carried out in the same way.
	
	To construct a digital map for positioning, the ground truth coordinates of the square structures need to be attained. For an indoor parking lot, we can refer to the layout design, where the exact position of every pillar is explicit. For an outdoor parking lot, it is also feasible to generate the map with the aid of high-precision GPS and a LiDAR, using the algorithm presented above. It is noted that the relative position of landmarks need to be different to some extent, so that they can be distinguished in localization stage. In the parking lot our robot to work in, the newly placed charging piles is designed non-equidistant. In addition, if the current environment of a parking lot is not sufficient to meet the positioning requirements, some square pillars can be set artificially as landmarks for robot localization, such as square pillars hanging from the ceiling or standing on the ground.

\section{Localization}
The objective of the localization of the parking robot is to deduce a continuous and precise estimate of its global pose in the parking lot. It is inevitable that all forms of perception systems have inherent measurement inaccuracies, the accurate state of the observed system could not directly recreate by sensors including IMU, GPS or LiDAR, especially when low cost is one of the system design goals. Hence, it is indispensable to employ stochastic approaches to generate precise pose approximation of automated robot. In our work, Monte Carlo Localization is implemented, in which a set of $N$ particles from time k $P_k=[p_k^1,...,p_k^N]$ and their associated weights $W_k=[w_k^1,...,w_k^N]$ are employed to maintain the pose estimate. The detailed implementation of the filter and how to associate it with the previous extracted corners are shown as follows.
	
	\subsection{Particle State and Initialization}
	The state of each particle is represented by Universal Transverse Mercator (UTM) coordinates, which comprises three components: easting $E$, Northing $N$, and an orientation angle towards the east axis $\psi$. If the application scenario is limited in a specific parking lot, we can narrow down the scope by setting the reference frame to a similar but smaller UTM cell using the same coordinate definition.  Therefore, the state vector of particle $i$ at time step $k$ is defined as  
	$$p_k^i=[E\  N\  \psi]^T $$
	To perform Monte Carlo Localization, all $N$ particles should be distributed around the configuration space within the parking lot. For an outdoor parking lot, a normal GPS receiver give the estimate of the particle origin. While for an indoor parking lot, initialization areas are set to provide the initial position of particles, due to the absence of GPS signal indoor. 
	\iffalse
	It is also feasible to skip the initialization step in our approach as long as we set the original pose uncertainty to a large value corresponding to the size of the parking lot, however, a rough initialization is recommended so that the reliability and quickness of the localization can be guaranteed at the beginning.\fi
	With an initial pose estimate, $N$ particles are generated, the pose of which is subject to normal distributions. The variance of the normal distribution is determined by the uncertainty of the initial pose.
	
	\subsection{Motion Update}
	Before an update of the particle weights is performed, the hypothesized robot state should be predicted to account for the motion of the robot within a time span. The odometry or IMU equipped on the parking robot can measure the speed $\widetilde{v}_k$ and yaw rate $\widetilde{\dot{\psi}}_k$ of the robot, afterwards prediction over $\Delta t$ is carried out using Bicycle Model. The well known dead reckoning formulation is shown as follows:
	
	\begin{align}
	&	\Delta\psi_k=\widetilde{\dot{\psi}}_k\cdot\Delta t  \\
	&	\psi_{k+1}=\psi_k+\Delta\psi_k \\
	&	E_{k+1}=E_k+\frac{\widetilde{v}_k}{\widetilde{\dot{\psi}}_k} \cdot  [sin(\psi_k+\Delta\psi_k)-sin(\psi_k)] \\
	&	N_{k+1}=N_k+\frac{\widetilde{v}_k}{\widetilde{\dot{\psi}}_k} \cdot  [cos(\psi_k)-cos(\psi_k+\Delta\psi_k)]
	\end{align}

	\begin{figure}
		\centering
		\subfloat[Corresponding scene in V-REP at the moment]{\includegraphics[width= 3.3in]{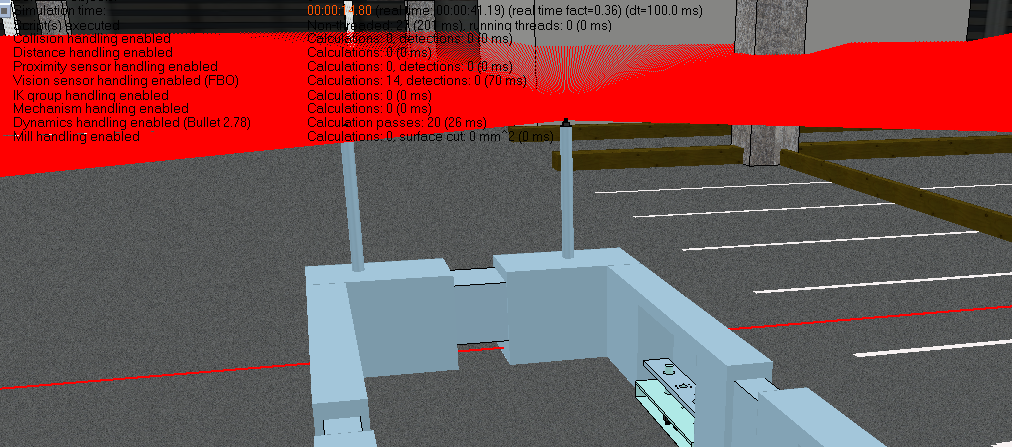}}\\
		\subfloat[L-Shape Fitting(left) and measurement update(right)]{\includegraphics[width= 3.3in]{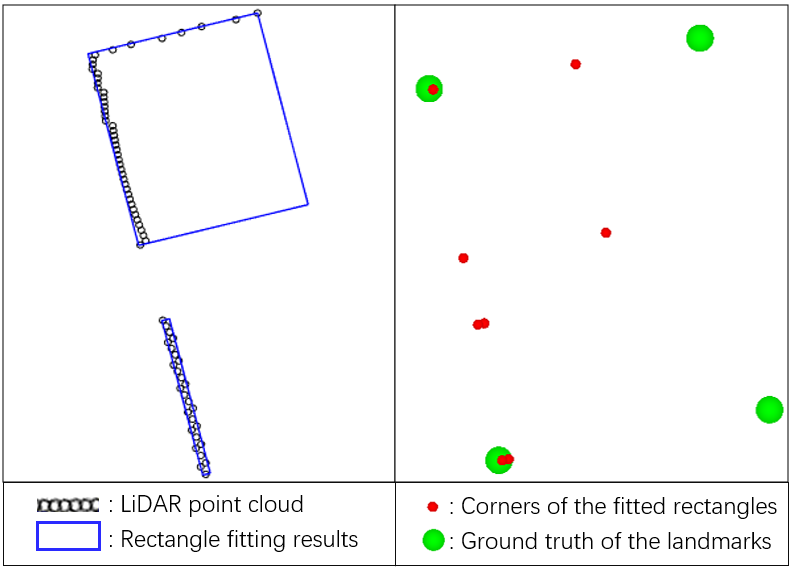}}
		\caption{An example of filtering process (best viewed in color)}
		\label{figs:point}
	\end{figure}
	
	\subsection{Measurement Update}
	Whenever a frame of LiDAR data is obtained, we merge it with the previous few frames to enhance the performance of the square landmark extraction mentioned in section IV. After the extraction approach is carried out, a vector of corner coordinates $[x_1,y_1,...,x_m,y_m] $ is acquired as the observed landmarks. Then the observed landmarks are associated with the truth value stored in the off-line map. At this stage, some filtering algorithms are performed to exclude unreliable pairs. According to the matching result, the Euclidean distance between the landmark coordinates and particles will be calculated and determine the updated weight of each particle.

		\subsubsection{Landmark Association}
		In order to associate the observed landmarks with the digital map, coordinate system conversion needed to be performed firstly. For each particle $p_k^i=[E,N,\psi]^T $ , we can easily convert the given observation coordinates $[x,y]^T$ from robot to map coordinate system:
		
		\begin{equation}									
			\dbinom{x_{conv}}{y_{conv}} = 
			\begin{pmatrix}
			cos\psi & -sin\psi \\
			sin\psi & cos\psi\\
			\end{pmatrix} 
			\cdot \dbinom{x}{y} + \dbinom{E}{N}
		\end{equation}
		
		For each converted observation coordinate $[x_{conv}\  y_{conv}]^T$, the closest landmark coordinate in digital map is to be found. Here we apply the 2-D Nearest Neighbor Search based on KD-Tree. In the first place, a KD-Tree is constructed with the converted observation. Thereafter, ground truth landmarks will be selected from digital map according to the previous pose estimation and the given LiDAR range. We then traverse all the selected landmarks(truth-value) to find the nearest node in KD-Tree(observation). In the end, each cloest pair will share a same landmark ID. 
		
		To filter out wrong matches, the Euclidean distance of all the pairing landmarks have to be less than a threshold. In this way, those corners that are incorrectly detected during the Landmark perception are excluded. In addition, every selected landmark will not associate with more than one observation, considering the scale requirement for landmarks when constructing the digital map. 
		%the landmark corners that are too close are already removed when constructing the digital map.
		
		Here is an example to illustrate this filtering process. The bottom left picture in Fig.~\ref{figs:point} shows the LiDAR point cloud and its rectangle fitting result. And the right one tells the groud truth corner position(in green) and estimate position(in red). The big rectangle is actually a stairwell in the parking lot. This situation occurs when the robot senses part of stairwell's outer wall, but a pillar happens to stand between the robot and the wall, which blocks the detection of laser, resulting in a partial gap in the point cloud of the wall(shown in Fig.~\ref{figs:point}(a)). In this case, the L-Shape fitting method will get two rectangles, which means 8 corners are included in the observation set. After the filtering process, however, only the two points closest to the top left and bottom left are retained due to the distance threshold(corresponding to the radius of the green marks in Fig.~\ref{figs:point}(b)). There might be two points (in the bottom left) meet the distance requirement simultaneously, but only the closest one left. Consequently this landmark won't be counted twice in the weight update step.

		\subsubsection{Particle Weight Update}
		In this stage, we calculate how well the set of converted observations $\mathcal{Z}_k =\{z^1,...,z^{Z_k}\} $ matches the stored landmark map $\mathcal{M}_k =\{m^1,...,m^{M_k}\} $ . It is noted that the association has been employed within these two sets. Let $d_{lat}$ and $d_{lon}$ represent the lateral and longitudinal Euclidean distance between the converted observations $\mathcal{Z}_k$ and landmarks $\mathcal{M}_k$ respectively, $\sigma_{lat}$ and $\sigma_{lon}$ be the uncertainty of Landmark measurement in different directions. Similar to the approach in~\cite{11}, the likelihood of all combinations of transformed observations $z^j$ and landmark $m_l$ is calculated as follows:
		
		 \begin{align}
		 &g(z^j|m^l) = \frac{1}{2\pi\sigma_{lat}\sigma_{lon}} \cdot exp(-\frac{1}{2}\cdot\Gamma) 
		 \end{align}
		 with:
		 \begin{align}
		 &\Gamma = \frac{d_{lat}^2}{\sigma_{lat}^2} + \frac{d_{lon}^2}{\sigma_{lon}^2} 
		 \end{align}
		The weight of particle $p_k^j$ is now defined as:
		\begin{align}
		&w_k^i =\prod_{l=1}^{M_k} g(z^j|m^l)
		\end{align}
		
		In this way, each particle acquires its weight, which virtually indicates the reliability of the particle pose and serves as the basis of resampling.
		
		\begin{figure}
			\centering
			\subfloat[Top view]{\includegraphics[width= 3.3in]{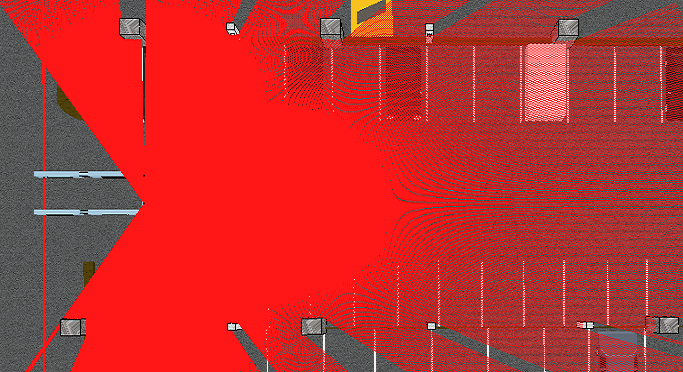}}\\
			\subfloat[Closer view]{\includegraphics[width= 3.3in]{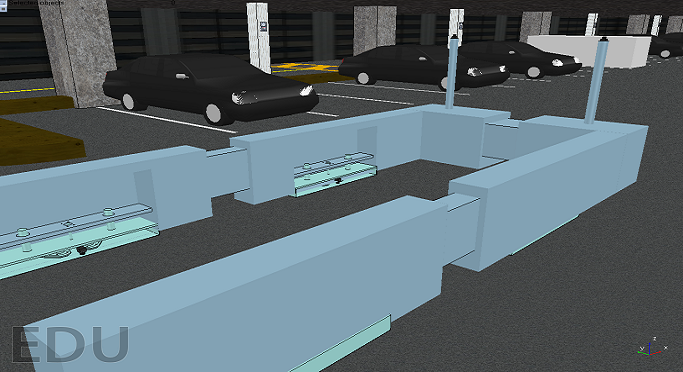}}
			\caption{Simulation environment}
			\label{figs:simulation}
		\end{figure}
	 \begin{table}
		\centering
		\caption{Parameters of the LiDAR}
		\label{table:lidar}
		\begin{tabular}{cc}
			\midrule
			Detection Range   &30m \& 270° \\
			Accuracy &$\pm50mm$\\
			Angular Resolution		&0.25° \\
			\bottomrule	
		\end{tabular}       
	\end{table}
	\subsection{Resampling and State Estimation}
	A discrete probability distribution is generated according to the weight vector of particles. Based on distribution, random integer $i$ on the interval $[1,N]$ is selected, representing the $i^{th}$ particle, and the probability of each individual integer to be selected is defined as: 
	\begin{align}
	&Prob (i) = \frac{w_k^i}{\sum_{i=1}^N w_k^i}
	\end{align}
	
	Once a random integer is produced, a newly particle is added to the resampled particle list. After iterations for N times, a new particle list is generated according to weight. In this way, the particles is moving in the direction of increasing weight during every measurement step. At the last of each measurement step, we select the best particle with highest weight to estimate the robot pose. Thus the self-localization of the robot is done.

\section{Evaluation}
In this section, localization system is tested in a simulation environment built on V-REP. The detailed configuration of the simulation environment and the employed sensors is illustrated in subsection A, after which the results of the test are shown and analyzed as well.

\subsection{Simulation Environment}
	An indoor parking lot is built in robot simulator V-REP, with 23 pillars and 18 charging piles in it. In order to get closer to the real parking scene, the layout of the parking lot is designed based on an existing parking lot with a number of parked vehicles in it. Fig. ~\ref{figs:simulation} shows the simulation environment.
	
	All sensor data are provided by the simulation environment. For position initialization, it offers a initial position with noise, the accuracy of which is 5 meters for the position and 2° for the heading. IMU data is updated at 100Hz, while LiDAR data at 5Hz. The parameters of the LiDAR are presented in Table.~\ref{table:lidar}.

\subsection{Experimental Results}
\begin{figure}
		\centering
		\subfloat{\includegraphics[width= 3.3in]{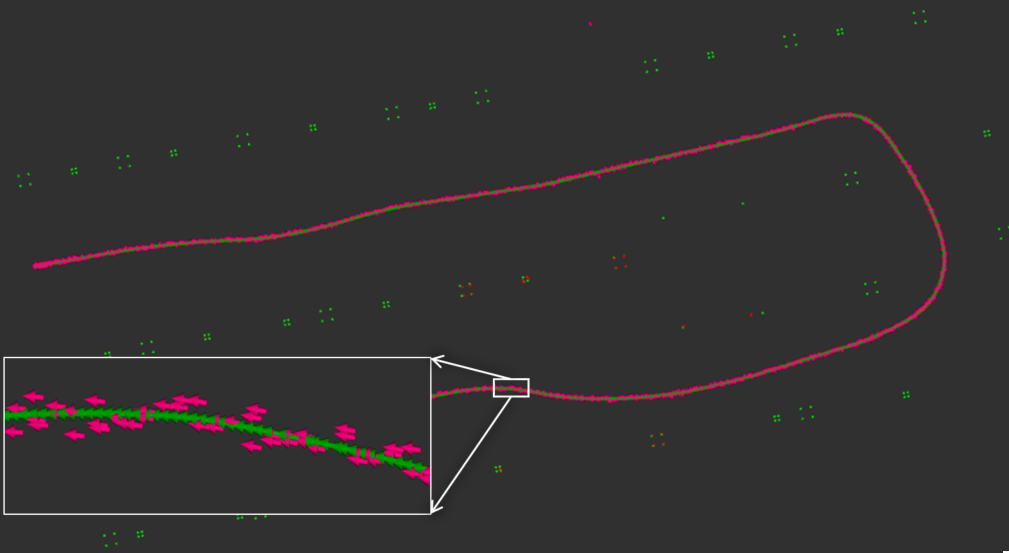}}
		\caption{The trajectory of estimate in pink and ground truth in green (best viewed in color)}
		\label{figs:trajectory}
	\end{figure}

\begin{figure}
		\centering
		\subfloat[Longitudinal Error]{\includegraphics[width = 3.2in]{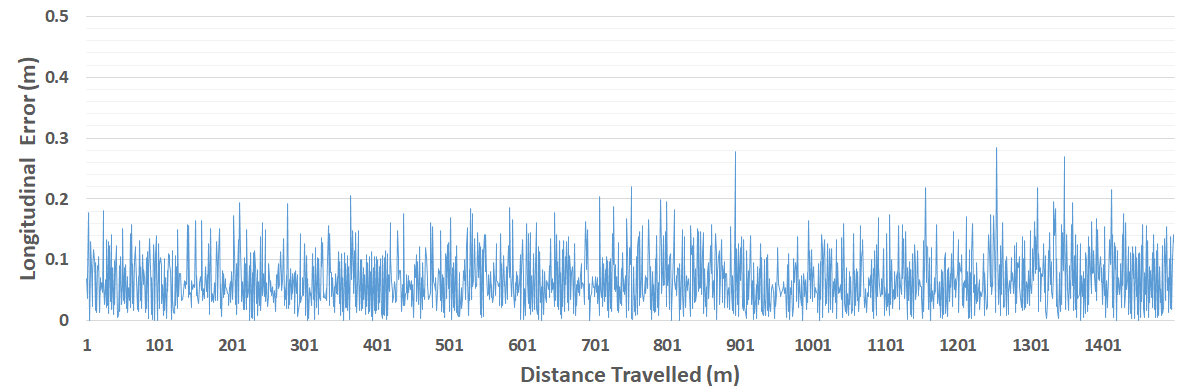}}\\
		\subfloat[Lateral Error]{\includegraphics[width = 3.2in]{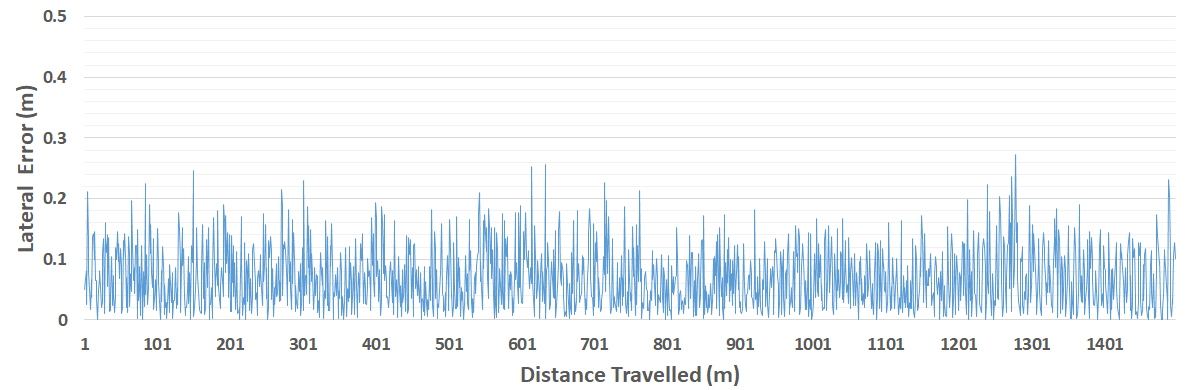}}\\
		\subfloat[Heading Error]{\includegraphics[width = 3.2in]{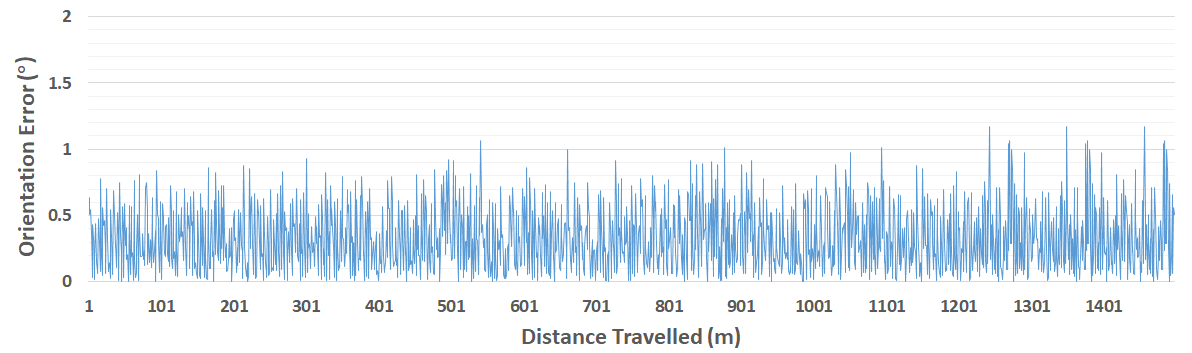}}
		\captionsetup{font={small}}
		\caption{Estimate error during the whole drive}
		\label{figs:error}
	\end{figure}

	The estimate trajectory and the ground truth route are showed in Fig.~\ref{figs:trajectory}. The total length of the driving route measures 1500 meters. In Fig.~\ref{figs:error}, the error for position as well as heading during the entire path is shown. The average lateral, longitudinal and orientation errors are listed in Table.~\ref{table:average error}. It is evident, that the localization using square landmarks is able to maintain position error below 0.2 m and an orientation error less than 1$^{\circ}$  during the whole drive.

	\begin{table}
		\centering
		\caption{Average error in the whole drive}
		\label{table:average error}
		\begin{tabular}{cc}
			\toprule
			Direction   &Error \\
			\midrule
			Longitudinal &0.098 m\\
			Lateral		&0.085 m \\
			Orientation		&0.46$^{\circ}$ \\
			\bottomrule	
		\end{tabular}       
	\end{table}

\section{Conclusion}
In this paper, a localization approach based on square-like landmark using a digital map is presented. We demonstrate that square-like structures can serve as reliable landmarks in a parking lot for autonomous parking robot cars. The realized accuracy is sufficient for the parking robot to locate itself and carry out its task. The next step is to conduct a field test of the approach with our parking robot. Applying this method to a commercial vehicle is a future goal. It is also a practical idea to combine other landmarks including poles, lanes or pavements with square landmarks, making it a commonly applicable localization method.
\bibliographystyle{ieeetr}
\bibliography{Lshape}

\end{document}